\pdfoutput=1

\documentclass[11pt]{article}

\usepackage[]{ACL2023}

\usepackage{times}
\usepackage{latexsym}

\usepackage[T1]{fontenc}

\usepackage[utf8]{inputenc}

\usepackage{microtype}

\usepackage{xcolor}

\usepackage{fancyvrb}
\usepackage{bm}
\usepackage{amssymb}
\usepackage{amsmath}
\usepackage{graphicx}
\usepackage{adjustbox}
\usepackage{booktabs}
\usepackage{multirow}
\usepackage{tabularx}
\usepackage{inconsolata}

\title{PRiSM: Enhancing Low-Resource Document-Level Relation Extraction with Relation-Aware Score Calibration}

\author{Minseok Choi \hspace{0.5cm} Hyesu Lim \hspace{0.5cm} Jaegul Choo \\
  KAIST AI \\
  \texttt{\{minseok.choi, hyesulim, jchoo\}@kaist.ac.kr}
}

\begin{document}
\maketitle

\begin{abstract}
Document-level relation extraction (DocRE) aims to extract relations of all entity pairs in a document.
A key challenge in DocRE is the cost of annotating such data which requires intensive human effort.
Thus, we investigate the case of DocRE in a low-resource setting, and we find that existing models trained on low data overestimate the \verb|NA| (``no relation'') label, causing limited performance.
In this work, we approach the problem from a calibration perspective and propose PRiSM, which learns to adapt logits based on relation semantic information.
We evaluate our method on three DocRE datasets and demonstrate that integrating existing models with PRiSM improves performance by as much as 26.38 F1 score, while the calibration error drops as much as 36 times when trained with about 3\% of data.
The code is publicly available at \url{https://github.com/brightjade/PRiSM}.
\end{abstract}
\section{Introduction}

Document-level relation extraction (DocRE) is a fundamental task in natural language understanding, which aims to identify relations between entities that exist in a document.
A major challenge in DocRE is the cost of annotating such documents, requiring annotators to consider relations of all possible entity combinations~\cite{yao2019docred, zaporojets2021dwie, tan2022redocred}.
However, there is a lack of ongoing studies investigating the low-resource setting in DocRE~\cite{zhou2022continual}, and we discover that most of the current DocRE models show subpar performance when trained with a small set of data.
We argue that the reason is two-fold. First, the long-tailed distribution of DocRE data encourages models to be overly confident in predicting frequent relations and less sure about infrequent ones~\cite{du2022eracl, tan2022kd}.
Out of the 96 relations in DocRED~\cite{yao2019docred}, a widely-used DocRE dataset, the 7 most frequent relations account for 55\% of the total relation triples.
Under the low-resource setting, chances to observe infrequent relations become much harder.
Second, DocRE models predict the \verb|NA| (``no relation'') label if an entity pair does not express any relation.
In DocRED, about 97\% of all entity pairs have the \verb|NA| label.
With limited data, there is a much less signal for ground-truth (GT) labels during training, resulting in models overpredicting the \verb|NA| label instead.

High confidence in common relations and the \verb|NA| label and low confidence in rare relations suggest that models may be miscalibrated.
We hypothesize that lowering the former and raising the latter would improve the overall RE performance.
At a high level, we wish to penalize logits of frequent labels (including \verb|NA|) and supplement logits of infrequent labels such that models are able to predict them without seeing them much during training.
To implement such behavior, we leverage relation semantic information, which has proved to be effective in low-resource sentence-level RE~\cite{yang2020enhance, dong2021mapre, zhang2022better}.

\begin{figure}
    \centering
    \includegraphics[width=\linewidth]{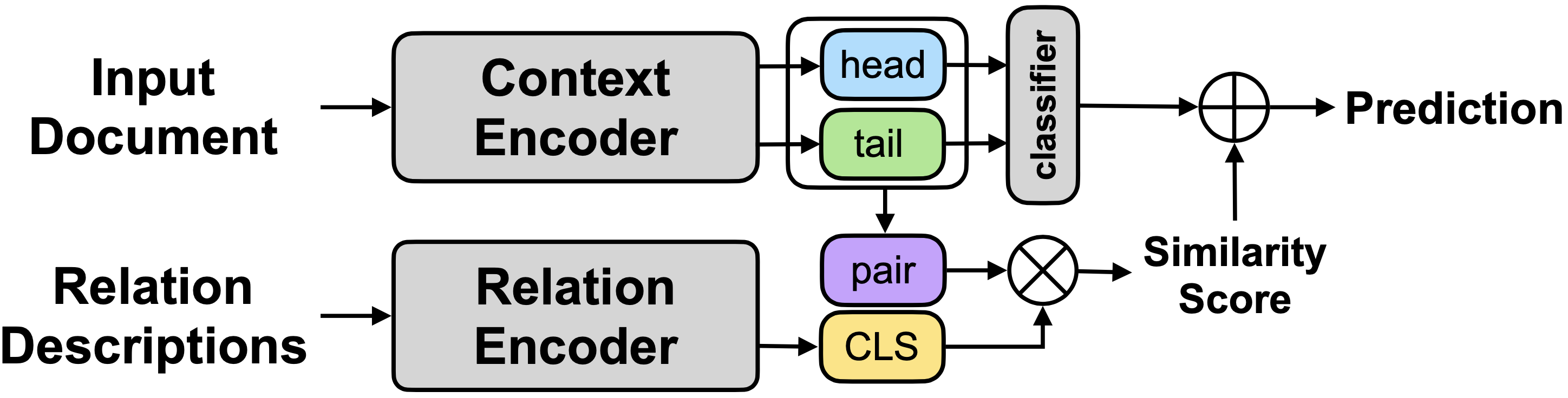}
    \caption{An overview of our proposed method. Top represents the original DocRE framework. PRiSM (bottom) leverages relation descriptions to compute scores for each relation triple. These scores are then used to reweight the prediction logits.}
    \label{fig:1}
\end{figure}

In this work, we propose the \textbf{P}air-\textbf{R}elat\textbf{i}on \textbf{S}imilarity \textbf{M}odule (PRiSM) that learns to adapt logits by exploiting semantic information from label descriptions, as depicted in Figure~\ref{fig:1}.
Specifically, we compute a similarity function for each entity pair embedding, constructed from two entities of interest, with relation embeddings, built from corresponding label descriptions.
PRiSM then learns relation representations to output adaptive scores for each relation triple.
Note that previous work mostly utilized relation representations for self-supervised learning~\cite{dong2021mapre, du2022eracl, zhou2022continual}, whereas PRiSM uses them to directly adjust logits, which brings a calibration effect.
To elaborate further, let us say that classification logits are statistical scores and similarities are semantic scores. We have four scenarios: 1) relation is common and GT, 2) relation is common but not GT, 3) relation is uncommon but GT, and 4) relation is uncommon and not GT.
In Cases 1 and 4, both statistical and semantic scores are either high or low, and thus, appending PRiSM mostly would not affect the original RE predictions.
In Case 2, the statistical score is high, but the semantic score is low, possibly negative to penalize the statistical score. This is the case of PRiSM decreasing the confidence of common relations and \verb|NA| label.
In Case 3, the statistical score is low, but the semantic score is high, which is the case of PRiSM increasing the confidence of uncommon relations.
As such, PRiSM incorporates both statistical and semantic scores such that the confidence is adjusted regardless of the relation frequency.

Our technical contributions are three-fold.
First, we propose PRiSM, a relation-aware calibration technique that improves model performance and adjusts model confidence on low-resource DocRE. Second, we demonstrate the performance improvement across various state-of-the-art models integrated with PRiSM. Third, we validate the effectiveness of our method on widely-used long-tailed DocRE datasets and calibration metrics.

\section{Methodology} \label{sec:2}

\subsection{Problem Formulation} \label{ssec:2.1}
Given a document $d$, a set of $n$ annotated entities $\mathcal{E}=\{e_i\}_{i=1}^n$, and a pre-defined set of relations $\mathcal{R} \cup \{$\Verb|NA|$\}$, the task of DocRE is to extract the relation triple set $\{(e_h, r, e_t)|e_h \in \mathcal{E}, r \in \mathcal{R}, e_t \in \mathcal{E}\} \subseteq \mathcal{E} \times \mathcal{R} \times \mathcal{E}$ from all possible relation triples, where $(e_h, r, e_t)$ denotes that a relation $r$ holds between head entity $e_h$ and tail entity $e_t$.
An entity $e_i$ may appear $k$ times in the document in which we denote corresponding instances as entity mentions $\{m_{ij}\}_{j=1}^{k}$. A relation $r$ exists between an entity pair $(e_h, e_t)$ if any pair of their mentions express the relation, and if they do not express any relation, the entity pair is then labeled as \Verb|NA|.

\subsection{Document-Level Relation Extraction} \label{ssec:2.2}
Given a document $d$ as an input token sequence $\bm{x} = [x_t]_{t=1}^l$, where $l$ is the length of the token sequence, we explicitly locate the position of entity mentions by inserting a special token ``*'' before and after each mention. The presence of the entity marker has proved to be effective from previous studies~\citep{zhang2017position, shi2019simple, soares2019matching}. The entity-marked document is then fed into a pre-trained language model (PLM) encoder, which outputs the contextual embeddings: $[\bm{h}_1, \bm{h}_2, ..., \bm{h}_l] = \text{Encoder}(\bm{x})$.
We take the embedding of ``*'' at the start of each mention as its mention-level representation $\bm{h}_{m_{ij}}$ of the entity $e_i$. For extracting the entity-level representation, we apply the logsumexp pooling over all mentions $\{m_{ij}\}_{j=1}^k$ of the entity $e_i$:
\begin{equation} \label{eq:1}
    \bm{h}_{e_i} = \log \sum\limits_{j=1}^k \exp \left(\bm{h}_{m_{ij}}\right).
\end{equation}
The logsumexp pooling is a smooth version of max pooling and has been shown to accumulate weak signals from each different mention representation, which results in a better performance~\citep{jia2019document}.
We pass the embeddings of head and tail entities through a linear layer followed by non-linear activation to obtain the hidden representations: $\bm{z}_h = \tanh(\bm{W}_h \bm{h}_{e_h} + \bm{b}_h)$ and $\bm{z}_t = \tanh(\bm{W}_t \bm{h}_{e_t} + \bm{b}_t)$, where $\bm{W}_h, \bm{W}_t, \bm{b}_h, \bm{b}_t$ are learnable parameters.
Then we calculate a score for relation $r$ between entities $h$ and $t$ by taking a bilinear function:
\begin{equation} \label{eq:2}
    s_{(h,r,t)} = \bm{z}_h^\top \bm{W}_r \bm{z}_t + \bm{b}_r,
\end{equation}
where $\bm{W}_r, \bm{b}_r$ are learnable parameters.

\begin{table*}[h]
    \centering
    \begin{adjustbox}{width=\textwidth}
    {\small
        \begin{tabular}{ l c c c c | c c c c }
        \noalign{\hrule height 1pt}
                       & \multicolumn{4}{c|}{\textbf{DocRED}} & \multicolumn{4}{c}{\textbf{Re-DocRED}}  \\
                       & \multicolumn{2}{c}{\textbf{Dev}} & \multicolumn{2}{c|}{\textbf{Test}} & \multicolumn{2}{c}{\textbf{Dev}} & \multicolumn{2}{c}{\textbf{Test}} \\
        \textbf{Model} & Ign $F_1$ & $F_1$ & Ign $F_1$ & $F_1$ & Ign $F_1$ & $F_1$ & Ign $F_1$ & $F_1$ \\
        \hline 
            \multicolumn{9}{l}{\textit{3\% training examples} $(N=100)$} \\
        \hline 
        BERT\textsubscript{BASE} & $10.27 \pm 1.82$  & $10.44 \pm 1.90$ & $11.36$ & $11.50$ & $28.65 \pm 2.87$  & $29.40 \pm 3.19$ & $28.77 \pm 3.34$ & $29.44 \pm 3.67$ \\
        BERT\textsubscript{BASE} + \textbf{PRiSM} & $\mathbf{35.06} \pm 0.94$ & $\mathbf{37.02} \pm 0.88$ & $\mathbf{35.79}$ & $\mathbf{37.88}$ & $\mathbf{47.39} \pm 0.79$ & $\mathbf{49.09} \pm 0.90$ & $\mathbf{46.90} \pm 1.59$ & $\mathbf{48.57} \pm 1.73$ \\
        RoBERTa\textsubscript{BASE} & $20.70 \pm 1.91$  & $21.31 \pm 1.87$ & $21.74$ & $22.25$ & $39.66 \pm 2.25$  & $40.74 \pm 1.89$ & $39.42 \pm 2.80$ & $40.53 \pm 2.43$ \\ 
        RoBERTa\textsubscript{BASE} + \textbf{PRiSM} & $\mathbf{32.40} \pm 0.85$ & $\mathbf{34.49} \pm 0.76$ & $\mathbf{32.20}$ & $\mathbf{34.32}$ & $\mathbf{47.71} \pm 1.03$ & $\mathbf{49.40} \pm 1.14$ & $\mathbf{47.31} \pm 0.96$ & $\mathbf{49.04} \pm 1.05$ \\
        SSAN-BERT\textsubscript{BASE} & $10.92 \pm 0.88$  & $11.18 \pm 0.89$ & $11.93$ & $12.16$ & $28.89 \pm 1.68$  & $29.01 \pm 1.69$ & $28.64 \pm 1.89$ & $29.29 \pm 1.94$ \\ 
        SSAN-BERT\textsubscript{BASE} + \textbf{PRiSM} & $\mathbf{32.86} \pm 2.35$ & $\mathbf{34.76} \pm 2.50$ & $\mathbf{34.00}$ & $\mathbf{36.03}$ & $\mathbf{46.49} \pm 1.16$ & $\mathbf{48.11} \pm 1.40$ & $\mathbf{46.51} \pm 1.77$ & $\mathbf{48.11} \pm 2.00$ \\
        ATLOP-BERT\textsubscript{BASE} & $38.99 \pm 2.30$  & $40.50 \pm 2.07$ & $40.88$ & $42.37$ & $49.45 \pm 2.09$  & $50.60 \pm 1.95$ & $49.24 \pm 2.25$ & $50.32 \pm 2.13$ \\
        ATLOP-BERT\textsubscript{BASE} + \textbf{PRiSM} & $\mathbf{40.59} \pm 0.68$ & $\mathbf{42.09} \pm 0.66$ & $\mathbf{40.94}$ & $\mathbf{42.43}$ & $\mathbf{50.10} \pm 0.53$ & $\mathbf{51.12} \pm 0.64$ & $\mathbf{50.15} \pm 1.11$ & $\mathbf{51.14} \pm 1.17$ \\

        \hline 
            \multicolumn{9}{l}{\textit{10\% training examples} $(N=305)$}  \\
        \hline 
        BERT\textsubscript{BASE} & $39.84 \pm 0.92$  & $41.55 \pm 0.99$ & $40.98$ & $42.98$ & $52.34 \pm 0.66$  & $53.54 \pm 0.80$ & $52.34 \pm 0.68$ & $53.54 \pm 0.84$ \\
        BERT\textsubscript{BASE} + \textbf{PRiSM} & $\mathbf{46.01} \pm 0.12$ & $\mathbf{48.02} \pm 0.13$ & $\mathbf{45.52}$ & $\mathbf{47.83}$ & $\mathbf{58.10} \pm 0.31$ & $\mathbf{59.86} \pm 0.27$ & $\mathbf{57.75} \pm 0.65$ & $\mathbf{59.53} \pm 0.51$ \\
        RoBERTa\textsubscript{BASE} & $43.42 \pm 1.09$  & $45.20 \pm 1.09$ & $43.78$ & $45.63$ & $54.82 \pm 1.85$  & $56.10 \pm 1.80$ & $55.36 \pm 2.18$ & $56.67 \pm 2.06$ \\
        RoBERTa\textsubscript{BASE} + \textbf{PRiSM} & $\mathbf{46.60} \pm 0.20$ & $\mathbf{48.57} \pm 0.29$ & $\mathbf{47.02}$ & $\mathbf{49.22}$ & $\mathbf{59.51} \pm 0.36$ & $\mathbf{61.19} \pm 0.32$ & $\mathbf{59.08} \pm 0.61$ & $\mathbf{60.80} \pm 0.52$ \\
        SSAN-BERT\textsubscript{BASE} & $40.00 \pm 1.62$ & $41.65 \pm 1.63$ & $41.11$ & $43.03$ & $53.57 \pm 0.83$  & $54.86 \pm 0.81$ & $53.67 \pm 1.55$ & $54.94 \pm 1.52$ \\
        SSAN-BERT\textsubscript{BASE} + \textbf{PRiSM} & $\mathbf{46.14} \pm 0.15$ & $\mathbf{48.18} \pm 0.09$ & $\mathbf{45.48}$ & $\mathbf{47.72}$ & $\mathbf{58.47} \pm 0.39$ & $\mathbf{60.17} \pm 0.36$ & $\mathbf{58.21} \pm 0.31$ & $\mathbf{59.93} \pm 0.19$ \\
        ATLOP-BERT\textsubscript{BASE} & $49.93 \pm 1.11$  & $51.61 \pm 1.16$ & $50.04$ & $51.85$ & $60.38 \pm 0.46$  & $61.52 \pm 0.29$ & $60.46 \pm 0.55$ & $61.54 \pm 0.29$ \\
        ATLOP-BERT\textsubscript{BASE} + \textbf{PRiSM} & $\mathbf{50.20} \pm 0.68$ & $\mathbf{51.83} \pm 0.64$ & $\mathbf{50.29}$ & $\mathbf{52.17}$ & $\mathbf{60.58} \pm 0.18$ & $\mathbf{61.68} \pm 0.17$ & $\mathbf{60.90} \pm 0.37$ & $\mathbf{61.97} \pm 0.40$ \\
        
        \hline 
            \multicolumn{9}{l}{\textit{100\% training examples} $(N=3053)$} \\
        \hline 
        BERT\textsubscript{BASE} & $57.15 \pm 0.17$  & $59.18 \pm 0.05$ & $57.02$ & $59.35$ & $71.70 \pm 0.61$  & $73.17 \pm 0.55$ & $71.01 \pm 0.88$ & $72.48 \pm 0.78$ \\
        BERT\textsubscript{BASE} + \textbf{PRiSM} & $\mathbf{57.82} \pm 0.10$ & $\mathbf{59.93} \pm 0.15$ & $\mathbf{57.17}$ & $\mathbf{59.52}$ & $\mathbf{72.92} \pm 0.07$ & $\mathbf{74.25} \pm 0.07$ & $\mathbf{72.35} \pm 0.07$ & $\mathbf{73.69} \pm 0.11$ \\
        RoBERTa\textsubscript{BASE} & $58.24 \pm 0.36$  & $60.19 \pm 0.38$ & $58.00$ & $60.10$ & $74.00 \pm 0.20$  & $75.20 \pm 0.20$ & $73.56 \pm 0.04$ & $74.75 \pm 0.04$ \\
        RoBERTa\textsubscript{BASE} + \textbf{PRiSM} & $\mathbf{58.73} \pm 0.09$ & $\mathbf{60.70} \pm 0.02$ & $\mathbf{58.36}$ & $\mathbf{60.51}$ & $\mathbf{74.50} \pm 0.09$ & $\mathbf{75.71} \pm 0.06$ & $\mathbf{74.17} \pm 0.10$ & $\mathbf{75.38} \pm 0.10$ \\
        SSAN-BERT\textsubscript{BASE} & $57.59 \pm 0.35$  & $59.62 \pm 0.24$ & $57.71$ & $59.79$ & $72.59 \pm 0.15$  & $74.01 \pm 0.15$ & $71.95 \pm 0.11$ & $73.37 \pm 0.11$ \\
        SSAN-BERT\textsubscript{BASE} + \textbf{PRiSM} & $\mathbf{58.20} \pm 0.20$ & $\mathbf{60.27} \pm 0.14$ & $\mathbf{58.02}$ & $\mathbf{60.27}$ & $\mathbf{73.22} \pm 0.10$ & $\mathbf{74.65} \pm 0.07$ & $\mathbf{72.37} \pm 0.19$ & $\mathbf{73.80} \pm 0.18$ \\
        ATLOP-BERT\textsubscript{BASE} & $59.22 \pm 0.17$  & $61.18 \pm 0.10$ & $\mathbf{58.99}$ & $\mathbf{61.08}$ & $72.78 \pm 0.46$  & $73.73 \pm 0.37$ & $72.60 \pm 0.41$ & $73.51 \pm 0.38$ \\
        ATLOP-BERT\textsubscript{BASE} + \textbf{PRiSM} & $\mathbf{59.51} \pm 0.09$ & $\mathbf{61.31} \pm 0.05$ & $58.80$ & $60.77$ & $\mathbf{72.85} \pm 0.29$ & $\mathbf{73.80} \pm 0.35$ & $\mathbf{72.61} \pm 0.59$ & $\mathbf{73.53} \pm 0.53$ \\

        \noalign{\hrule height 1pt}
        \end{tabular}
    }
    \end{adjustbox}
    \caption{Performance (\%) on DocRED and Re-DocRED. Better scores between with and without PRiSM are in bold. The test results for DocRED are obtained by submitting the best dev model predictions to CodaLab\protect\footnotemark.}
    \label{tab:main_table}
\end{table*}

\subsection{PRiSM}  \label{ssec:2.3}

\begin{table}[]
    \centering
    \begin{adjustbox}{width=\linewidth}
    {\small
        \begin{tabular}{ l c c c c }
        \noalign{\hrule height 1pt}
        \textbf{Model} & Macro & Macro@500 & Macro@200 & Macro@100 \\
        \hline 
            \multicolumn{3}{l}{$N=100$}  \\
        \hline 
        BERT\textsubscript{BASE}                     & $0.36 \pm 0.05$      & $0$       & $0$      & $-$ \\
        BERT\textsubscript{BASE} + \textbf{PRiSM}    & $\bm{7.77} \pm 1.87$ & $\bm{4.08} \pm 0.43$ & $\bm{0.44} \pm 0.33$ & $-$ \\
        RoBERTa\textsubscript{BASE}                  & $1.18 \pm 0.28$      & $0$       & $0$      & $-$ \\
        RoBERTa\textsubscript{BASE} + \textbf{PRiSM} & $\bm{6.41} \pm 0.77$ & $\bm{2.31} \pm 0.82$ & $\bm{0.38} \pm 0.28$ & $-$ \\
        \hline 
            \multicolumn{3}{l}{$N=305$}  \\
        \hline 
        BERT\textsubscript{BASE}                     & $9.31 \pm 1.59$      & $3.70 \pm 1.46$      & $0.29 \pm 0.17$      & $0$ \\
        BERT\textsubscript{BASE} + \textbf{PRiSM}    & $\bm{20.19} \pm 0.70$ & $\bm{14.91} \pm 0.64$ & $\bm{7.73} \pm 0.17$ & $\bm{2.19} \pm 1.16$ \\
        RoBERTa\textsubscript{BASE}                  & $14.80 \pm 0.51$      & $9.13 \pm 0.61$      & $3.74 \pm 0.35$      & $0.83 \pm 0.89$ \\
        RoBERTa\textsubscript{BASE} + \textbf{PRiSM} & $\bm{21.03} \pm 0.27$ & $\bm{15.69} \pm 0.48$ & $\bm{8.37} \pm 0.16$ & $\bm{2.63} \pm 1.25$ \\
        \hline 
            \multicolumn{3}{l}{$N=3053$}  \\
        \hline 
        BERT\textsubscript{BASE}                     & $38.31 \pm 0.39$      & $34.06 \pm 0.45$ & $26.07 \pm 0.72$ & $\bm{19.73} \pm 0.96$ \\
        BERT\textsubscript{BASE} + \textbf{PRiSM}    & $\bm{38.89} \pm 0.52$ & $\bm{34.57} \pm 0.59$ & $\bm{26.51} \pm 0.65$ & $19.57 \pm 0.71$ \\
        RoBERTa\textsubscript{BASE}                  & $38.67 \pm 1.12$      & $34.28 \pm 1.22$ & $26.14 \pm 1.44$ & $18.69 \pm 1.70$ \\
        RoBERTa\textsubscript{BASE} + \textbf{PRiSM} & $\bm{39.12} \pm 0.57$ & $\bm{34.72} \pm 0.69$ & $\bm{26.45} \pm 1.01$ & $\bm{19.23} \pm 1.55$\\
        \noalign{\hrule height 1pt}
        \end{tabular}
    }
    \end{adjustbox}
    \caption{Dev performance (\%) on low-frequency relations in DocRED. Test results cannot be reported because the labels are not accessible.}
    \label{tab:macro}
\end{table}

Following previous work~\cite{zhang2022better}, we feed relation descriptions to a PLM encoder to obtain the relation embedding $\bm{z}_r$ for relation $r$.
The details of the relation descriptions used can be found in Appendix~\ref{appendix:a4}.
We then construct the entity pair-level representation $\bm{z}_{(h,t)}$ by mapping the head and tail embeddings to a linear layer followed by non-linear activation: $\bm{z}_{(h,t)} = \tanh(\bm{W}_{(h,t)} [\bm{z}_h ; \bm{z}_t] + \bm{b}_{(h,t)})$, where $\bm{z}_h$, $\bm{z}_t$ are concatenated and $\bm{W}_{(h,t)}$, $\bm{b}_{(h,t)}$ are learnable parameters.
An adaptive score for relation $r$ between entities $h$ and $t$ is computed by taking a similarity function between the entity pair embedding and relation embedding: $s'_{(h,r,t)} = sim(\bm{z}_{(h,t)}, \bm{z}_r)$,
where $sim(\cdot)$ is cosine similarity.
Formally, the probability of relation $r$ between entities $h$ and $t$ is simply an addition of two scores followed by sigmoid activation:
\begin{equation} \label{eq:4}
    P(r \mid e_h,e_t) = \sigma (s_{(h,r,t)} + \lambda s'_{(h,r,t)}),
\end{equation}
where $\lambda$ is the scale factor.
Finally, we optimize our model with the binary cross-entropy (BCE) loss:
\begin{equation} \label{eq:5}
    \mathcal{L} = - \frac{1}{T} \sum_{<h,t>} \sum_{r} \verb|BCE|(P(r|e_h,e_t), \bar{y}_{(h,r,t)}),
\end{equation}
where $\bar{y}$ is the target label and $T$ is the total number of relation triples.


\section{Experiments}

\subsection{Dataset}
We evaluate our framework on three public DocRE datasets.
DocRED~\citep{yao2019docred} is a widely-used human-annotated DocRE dataset constructed from Wikipedia and Wikidata.
Re-DocRED~\citep{tan2022redocred} is a revised dataset from DocRED, addressing the incomplete annotation problem.
DWIE~\citep{zaporojets2021dwie} is a multi-task document-level information extraction dataset consisting of news articles collected from Deutsche Welle.
Dataset statistics are shown in Table~\ref{tab:data_stats}.

\footnotetext{\url{https://codalab.lisn.upsaclay.fr/}}

\begin{figure*}
    \centering
    \includegraphics[width=\textwidth]{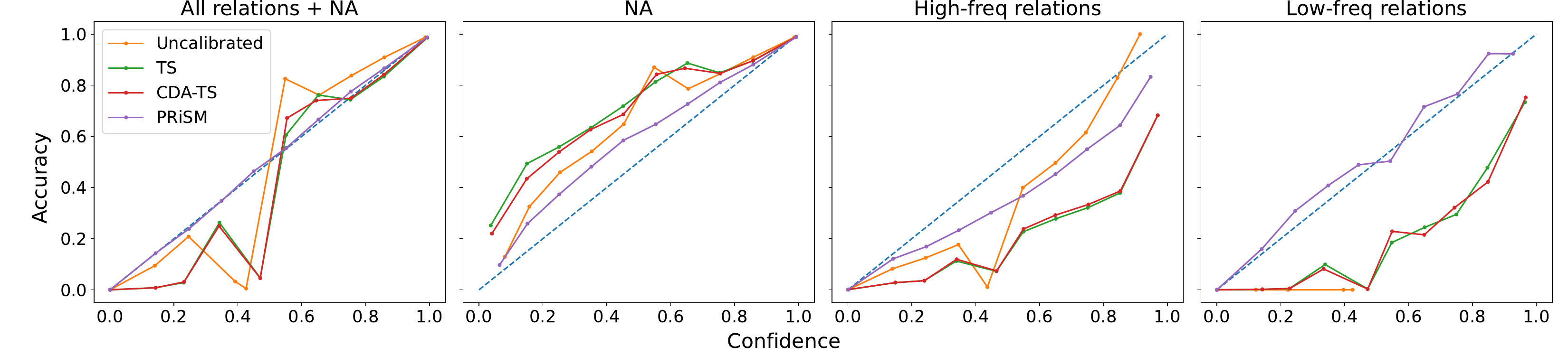}
    \caption{Reliability diagram for BERT\textsubscript{BASE} when trained with 3\% of DocRED data.}
    \label{fig:2}
\end{figure*}

\subsection{Implementation Details}
Our framework is built on PyTorch and Huggingface's Transformers library~\citep{wolf2019huggingface}. We use the cased BERT~\citep{devlin2018bert} and RoBERTa~\citep{liu2019roberta} for encoding the text and optimize their weights with AdamW~\citep{loshchilov2017decoupled}. We tune our hyperparameters to maximize the $F_1$ score on the development set.
The additional implementation details are included in Appendix~\ref{appendix:b}.
During inference, we predict all relation triples that have probabilities higher than the F1-maximizing threshold found in the development set.
We conduct our experiments with three different random seeds and report the averaged results.
Following~\citet{yao2019docred}, all models are evaluated on $F_1$ and Ign $F_1$, where Ign $F_1$ excludes the relations shared by the training and development/test sets.
Moreover, we measure \textbf{Macro}, which computes the average of per-class $F_1$, and \textbf{Macro@500}, \textbf{Macro@200}, and \textbf{Macro@100}, targeting rare relations where the frequency count in the training dataset is less than 500, 200, and 100, respectively.

\begin{table}[]
    \centering
    \begin{adjustbox}{width=\linewidth}
    {\small
        \begin{tabular}{ l c c c c }
        \noalign{\hrule height 1pt}
                       & \multicolumn{2}{c}{\textbf{Dev}} & \multicolumn{2}{c}{\textbf{Test}} \\
        \textbf{Model} & $F_1$ & Macro & $F_1$ & Macro \\
        \hline 
            \multicolumn{3}{l}{$N=100$}  \\
        \hline 
        BERT\textsubscript{BASE}                     & $11.97 \pm 1.78$      & $1.79 \pm 0.27$       & $12.41 \pm 2.29$      & $1.87 \pm 0.32$ \\
        BERT\textsubscript{BASE} + \textbf{PRiSM}    & $\bm{45.20} \pm 1.60$ & $\bm{10.34} \pm 1.91$ & $\bm{44.31} \pm 1.47$ & $\bm{9.47} \pm 1.46$ \\
        RoBERTa\textsubscript{BASE}                  & $50.27 \pm 1.57$      & $8.55 \pm 0.56$       & $48.29 \pm 1.74$      & $8.95 \pm 0.86$ \\
        RoBERTa\textsubscript{BASE} + \textbf{PRiSM} & $\bm{55.51} \pm 1.11$ & $\bm{12.76} \pm 2.03$ & $\bm{54.23} \pm 1.24$ & $\bm{13.80} \pm 0.64$ \\
        \hline 
            \multicolumn{3}{l}{$N=305$}  \\
        \hline 
        BERT\textsubscript{BASE}                     & $52.98 \pm 0.76$      & $15.68 \pm 1.71$      & $52.05 \pm 0.60$      & $14.80 \pm 0.63$ \\
        BERT\textsubscript{BASE} + \textbf{PRiSM}    & $\bm{58.23} \pm 0.40$ & $\bm{24.62} \pm 0.59$ & $\bm{57.05} \pm 0.23$ & $\bm{22.43} \pm 0.88$ \\
        RoBERTa\textsubscript{BASE}                  & $65.45 \pm 1.94$      & $21.72 \pm 1.29$      & $62.39 \pm 1.29$      & $20.39 \pm 0.76$ \\
        RoBERTa\textsubscript{BASE} + \textbf{PRiSM} & $\bm{71.18} \pm 1.98$ & $\bm{28.36} \pm 1.53$ & $\bm{67.12} \pm 2.02$ & $\bm{25.82} \pm 0.34$ \\
        \hline 
            \multicolumn{3}{l}{$N=587$}  \\
        \hline 
        BERT\textsubscript{BASE}                     & $62.06 \pm 0.33$      & $25.17 \pm 0.37$ & $60.78 \pm 0.25$ & $22.93 \pm 0.40$ \\
        BERT\textsubscript{BASE} + \textbf{PRiSM}    & $\bm{66.81} \pm 0.56$ & $\bm{28.17} \pm 0.54$ & $\bm{66.53} \pm 0.52$ & $\bm{29.31} \pm 1.13$ \\
        RoBERTa\textsubscript{BASE}                  & $76.23 \pm 0.72$      & $31.71 \pm 0.13$ & $74.07 \pm 0.77$ & $28.72 \pm 1.54$ \\
        RoBERTa\textsubscript{BASE} + \textbf{PRiSM} & $\bm{78.43} \pm 0.12$ & $\bm{32.85} \pm 0.37$ & $\bm{78.13} \pm 0.61$ & $\bm{33.66} \pm 1.24$\\
        \noalign{\hrule height 1pt}
        \end{tabular}
    }
    \end{adjustbox}
    \caption{Performance (\%) on the DWIE dataset.}
    \label{tab:dwie}
\end{table}

\subsection{Experimental Results}
To simulate the low-data setting, we reduce the number of training documents $N$ to 100 and 305, which is about 3\% and 10\% of the original data.
To create each of the settings, we repeat random sampling until the label distribution resembles that of the full data.
As shown in Table~\ref{tab:main_table}, we observe that performance increases consistently across different models when appended with PRiSM.
Particularly, PRiSM improves performance by a large margin when trained with just 3\% of data, as much as 24.43 Ign $F_1$ and 26.38 $F_1$ on the test set of DocRED for BERT\textsubscript{BASE}.
We also test PRiSM on RoBERTa\textsubscript{BASE} and two state-of-the-art models SSAN~\cite{xu2021ssan} and ATLOP~\cite{zhou2021atlop} and notice a similar trend, indicating that our method is effective on various existing models.
We additionally evaluate PRiSM using macro metrics in Table~\ref{tab:macro} and observe that adding PRiSM improves performance on infrequent relations, especially in the low-data setting.
Lastly, we validate our method on a different dataset DWIE, as illustrated in Table~\ref{tab:dwie}.

\subsection{Calibration Evaluation}

We measure model calibration on two metrics: expected calibration error (ECE)~\cite{naeini2015bayesian} and adaptive calibration error (ACE)~\cite{nixon2019ace}.
ECE partitions predictions into a fixed number of bins and computes a weighted average of the difference between accuracy and confidence over the bins, while ACE puts the same number of predictions in each bin.
We compare with general calibration methods such as temperature scaling (TS)~\cite{guo2017calibration} and class-distribution-aware TS (CDA-TS)~\cite{islam2021cdats}. As reported in Table~\ref{tab:calibration}, PRiSM outperforms other methods in both metrics, while also maintaining a comparable RE performance.
We also visualize with a reliability diagram~\cite{degroot1983comparison, niculescu2005predicting} in Figure~\ref{fig:2}.
We observe that PRiSM effectively lowers the confidence of the \verb|NA| label and raises the confidence of low-frequency relations (bottom 89). For high-frequency relations (top 7), confidence is adjusted in both ways. In any case, PRiSM displays the most stable, closest line to the perfect calibration (blue line).

\begin{table}[]
    \centering
    \begin{adjustbox}{width=\linewidth}
    {\small
        \begin{tabular}{ l | c c c | c c c }
            \noalign{\hrule height 1pt}
            & \multicolumn{3}{c|}{$N=100$} & \multicolumn{3}{c}{$N=305$} \\
            \hline
            \textbf{Method} & \bm{$F_1$}($\uparrow$) & \textbf{ECE}($\downarrow$) & \textbf{ACE}($\downarrow$) & \bm{$F_1$}($\uparrow$) & \textbf{ECE}($\downarrow$) & \textbf{ACE}($\downarrow$)\\
            \hline
            Uncalibrated & 10.82 & 0.359\% & 0.379\% & 42.56 & 0.137\% & 0.164\% \\
            TS & \textbf{38.19} & 0.144\% & 0.173\% & 48.49 & 0.053\% & 0.062\% \\
            CDA-TS & 37.82 & 0.139\% & 0.167\% & \textbf{48.54} & 0.057\% & 0.078\% \\
            PRiSM (ours) & 37.84 & \textbf{0.010\%} & \textbf{0.020\%} & 48.10 & \textbf{0.023\%} & \textbf{0.020\%} \\
            \noalign{\hrule height 1pt}
        \end{tabular}
    }
    \end{adjustbox}
    \caption{Comparison of calibration errors (with 10 bins) under a low-resource setting of DocRED.}
    \label{tab:calibration}
\end{table}

\section{Related Work}
With the introduction of DocRED~\citep{yao2019docred}, many approaches were proposed to extract relations from a document~\citep{wang2019twostep, ye2020coreferential, zhang2021docunet, xu2021ssan, zhou2021atlop, xie2022eider}. The long-tailed data problem of DocRE has been addressed in some studies~\citep{du2022eracl, tan2022kd}, as well as low-resource DocRE~\cite{zhou2022continual}; however, most require additional pretraining, which is compute- and cost-intensive, while PRiSM only requires adjusting logits in existing models. Low-resource RE has been extensively studied at the sentence level, and we specifically focus on leveraging label information~\cite{yang2020enhance, dong2021mapre, zhang2022better} in which PRiSM applies it to the document level.
In contrast to prior work in calibration~\cite{guo2017calibration, islam2021cdats}, our approach is relation-aware, updating logits at a much finer granularity.


\section{Conclusion and Future Work}
In this work, we propose a simple modular framework PRiSM, which exploits relation semantics to update logits.
We empirically demonstrate that our method effectively improves and calibrates DocRE models where the data is long-tailed and the \verb|NA| label is overestimated.
For future work, we can apply PRiSM to more tasks such as event extraction and dialogue state tracking, which also enclose long-tailed data and overestimation of ``null'' labels.

\section*{Limitations}
Although our approach is resilient to data scarcity, quite a few annotated documents are still required for the model to learn the pattern. The ultimate goal of DocRE is undoubtedly to build a model that is able to perform well on zero-shot, but we believe our approach takes a step toward that direction.
Moreover, we process the long documents (> 512 tokens) in a very naive way, as described in Appendix~\ref{appendix:a3}, and we think that exploration of long-sequence modeling on longer document data could further enrich the field of DocRE.

\section*{Acknowledgements}
This work was supported by Institute of Information \& communications Technology Planning \& Evaluation (IITP) grant funded by the Korea government (MSIT) (No.2019-0-00075, Artificial Intelligence Graduate School Program (KAIST)), the National Supercomputing Center with supercomputing resources including technical support (KSC-2022-CRE-0312), and Samsung Electronics Co., Ltd. We thank the anonymous reviewers for their constructive feedback.

\bibliography{anthology,custom}

\begin{thebibliography}{30}
\expandafter\ifx\csname natexlab\endcsname\relax\def\natexlab#1{#1}\fi

\bibitem[{Baldini~Soares et~al.(2019)Baldini~Soares, FitzGerald, Ling, and
  Kwiatkowski}]{soares2019matching}
Livio Baldini~Soares, Nicholas FitzGerald, Jeffrey Ling, and Tom Kwiatkowski.
  2019.
\newblock \href {https://doi.org/10.18653/v1/P19-1279} {Matching the blanks:
  Distributional similarity for relation learning}.
\newblock In \emph{Proceedings of the 57th Annual Meeting of the Association
  for Computational Linguistics}, pages 2895--2905, Florence, Italy.
  Association for Computational Linguistics.

\bibitem[{DeGroot and Fienberg(1983)}]{degroot1983comparison}
Morris~H DeGroot and Stephen~E Fienberg. 1983.
\newblock \href {https://www.jstor.org/stable/2987588} {The comparison and
  evaluation of forecasters}.
\newblock \emph{Journal of the Royal Statistical Society: Series D (The
  Statistician)}, 32(1-2):12--22.

\bibitem[{Devlin et~al.(2019)Devlin, Chang, Lee, and
  Toutanova}]{devlin2018bert}
Jacob Devlin, Ming-Wei Chang, Kenton Lee, and Kristina Toutanova. 2019.
\newblock \href {https://doi.org/10.18653/v1/N19-1423} {{BERT}: Pre-training of
  deep bidirectional transformers for language understanding}.
\newblock In \emph{Proceedings of the 2019 Conference of the North {A}merican
  Chapter of the Association for Computational Linguistics: Human Language
  Technologies, Volume 1 (Long and Short Papers)}, pages 4171--4186,
  Minneapolis, Minnesota. Association for Computational Linguistics.

\bibitem[{Dong et~al.(2021)Dong, Pan, and Luo}]{dong2021mapre}
Manqing Dong, Chunguang Pan, and Zhipeng Luo. 2021.
\newblock \href {https://doi.org/10.18653/v1/2021.emnlp-main.212} {{M}ap{RE}:
  An effective semantic mapping approach for low-resource relation extraction}.
\newblock In \emph{Proceedings of the 2021 Conference on Empirical Methods in
  Natural Language Processing}, pages 2694--2704, Online and Punta Cana,
  Dominican Republic. Association for Computational Linguistics.

\bibitem[{Du et~al.(2022)Du, Ma, Wu, Wu, Zhang, Long, and Ji}]{du2022eracl}
Yangkai Du, Tengfei Ma, Lingfei Wu, Yiming Wu, Xuhong Zhang, Bo~Long, and
  Shouling Ji. 2022.
\newblock \href {https://arxiv.org/abs/2205.10511} {Improving long tailed
  document-level relation extraction via easy relation augmentation and
  contrastive learning}.
\newblock \emph{arXiv preprint arXiv:2205.10511}.

\bibitem[{Guo et~al.(2017)Guo, Pleiss, Sun, and
  Weinberger}]{guo2017calibration}
Chuan Guo, Geoff Pleiss, Yu~Sun, and Kilian~Q Weinberger. 2017.
\newblock \href {https://arxiv.org/abs/1706.04599} {On calibration of modern
  neural networks}.
\newblock In \emph{International conference on machine learning}, pages
  1321--1330. PMLR.

\bibitem[{Islam et~al.(2021)Islam, Seenivasan, Ren, and
  Glocker}]{islam2021cdats}
Mobarakol Islam, Lalithkumar Seenivasan, Hongliang Ren, and Ben Glocker. 2021.
\newblock \href {https://arxiv.org/abs/2109.05263} {Class-distribution-aware
  calibration for long-tailed visual recognition}.
\newblock \emph{arXiv preprint arXiv:2109.05263}.

\bibitem[{Jia et~al.(2019)Jia, Wong, and Poon}]{jia2019document}
Robin Jia, Cliff Wong, and Hoifung Poon. 2019.
\newblock \href {https://doi.org/10.18653/v1/N19-1370} {Document-level n-ary
  relation extraction with multiscale representation learning}.
\newblock In \emph{Proceedings of the 2019 Conference of the North {A}merican
  Chapter of the Association for Computational Linguistics: Human Language
  Technologies, Volume 1 (Long and Short Papers)}, pages 3693--3704,
  Minneapolis, Minnesota. Association for Computational Linguistics.

\bibitem[{Liu et~al.(2019)Liu, Ott, Goyal, Du, Joshi, Chen, Levy, Lewis,
  Zettlemoyer, and Stoyanov}]{liu2019roberta}
Yinhan Liu, Myle Ott, Naman Goyal, Jingfei Du, Mandar Joshi, Danqi Chen, Omer
  Levy, Mike Lewis, Luke Zettlemoyer, and Veselin Stoyanov. 2019.
\newblock \href {https://arxiv.org/abs/1907.11692} {Roberta: A robustly
  optimized bert pretraining approach}.
\newblock \emph{arXiv preprint arXiv:1907.11692}.

\bibitem[{Loshchilov and Hutter(2019)}]{loshchilov2017decoupled}
Ilya Loshchilov and Frank Hutter. 2019.
\newblock \href {https://openreview.net/forum?id=Bkg6RiCqY7} {Decoupled weight
  decay regularization}.
\newblock In \emph{International Conference on Learning Representations}.

\bibitem[{Naeini et~al.(2015)Naeini, Cooper, and
  Hauskrecht}]{naeini2015bayesian}
Mahdi~Pakdaman Naeini, Gregory Cooper, and Milos Hauskrecht. 2015.
\newblock \href {https://ojs.aaai.org/index.php/AAAI/article/view/9602}
  {Obtaining well calibrated probabilities using bayesian binning}.
\newblock In \emph{Proceedings of the AAAI conference on artificial
  intelligence}, volume~29.

\bibitem[{Niculescu-Mizil and Caruana(2005)}]{niculescu2005predicting}
Alexandru Niculescu-Mizil and Rich Caruana. 2005.
\newblock \href
  {https://www.cs.cornell.edu/~alexn/papers/calibration.icml05.crc.rev3.pdf}
  {Predicting good probabilities with supervised learning}.
\newblock In \emph{Proceedings of the 22nd international conference on Machine
  learning}, pages 625--632.

\bibitem[{Nixon et~al.(2019)Nixon, Dusenberry, Zhang, Jerfel, and
  Tran}]{nixon2019ace}
Jeremy Nixon, Michael~W Dusenberry, Linchuan Zhang, Ghassen Jerfel, and Dustin
  Tran. 2019.
\newblock \href
  {https://openaccess.thecvf.com/content_CVPRW_2019/html/Uncertainty_and_Robustness_in_Deep_Visual_Learning/Nixon_Measuring_Calibration_in_Deep_Learning_CVPRW_2019_paper.html}
  {Measuring calibration in deep learning.}
\newblock In \emph{CVPR workshops}, volume~2.

\bibitem[{OpenAI(2023)}]{openai2023chatgpt}
OpenAI. 2023.
\newblock \href {https://chat.openai.com/} {Chat{GPT} {M}arch 23 {V}ersion}.

\bibitem[{Shi and Lin(2019)}]{shi2019simple}
Peng Shi and Jimmy Lin. 2019.
\newblock \href {https://arxiv.org/abs/1904.05255} {Simple bert models for
  relation extraction and semantic role labeling}.
\newblock \emph{arXiv preprint arXiv:1904.05255}.

\bibitem[{Tan et~al.(2022{\natexlab{a}})Tan, He, Bing, and Ng}]{tan2022kd}
Qingyu Tan, Ruidan He, Lidong Bing, and Hwee~Tou Ng. 2022{\natexlab{a}}.
\newblock \href {https://doi.org/10.18653/v1/2022.findings-acl.132}
  {Document-level relation extraction with adaptive focal loss and knowledge
  distillation}.
\newblock In \emph{Findings of the Association for Computational Linguistics:
  ACL 2022}, pages 1672--1681, Dublin, Ireland. Association for Computational
  Linguistics.

\bibitem[{Tan et~al.(2022{\natexlab{b}})Tan, Xu, Bing, Ng, and
  Aljunied}]{tan2022redocred}
Qingyu Tan, Lu~Xu, Lidong Bing, Hwee~Tou Ng, and Sharifah~Mahani Aljunied.
  2022{\natexlab{b}}.
\newblock \href {https://aclanthology.org/2022.emnlp-main.580} {Revisiting
  {D}oc{RED} - addressing the false negative problem in relation extraction}.
\newblock In \emph{Proceedings of the 2022 Conference on Empirical Methods in
  Natural Language Processing}, pages 8472--8487, Abu Dhabi, United Arab
  Emirates. Association for Computational Linguistics.

\bibitem[{Wang et~al.(2019)Wang, Focke, Sylvester, Mishra, and
  Wang}]{wang2019twostep}
Hong Wang, Christfried Focke, Rob Sylvester, Nilesh Mishra, and William Wang.
  2019.
\newblock \href {https://arxiv.org/abs/1909.11898} {Fine-tune bert for docred
  with two-step process}.
\newblock \emph{arXiv preprint arXiv:1909.11898}.

\bibitem[{Wolf et~al.(2020)Wolf, Debut, Sanh, Chaumond, Delangue, Moi, Cistac,
  Rault, Louf, Funtowicz, Davison, Shleifer, von Platen, Ma, Jernite, Plu, Xu,
  Le~Scao, Gugger, Drame, Lhoest, and Rush}]{wolf2019huggingface}
Thomas Wolf, Lysandre Debut, Victor Sanh, Julien Chaumond, Clement Delangue,
  Anthony Moi, Pierric Cistac, Tim Rault, Remi Louf, Morgan Funtowicz, Joe
  Davison, Sam Shleifer, Patrick von Platen, Clara Ma, Yacine Jernite, Julien
  Plu, Canwen Xu, Teven Le~Scao, Sylvain Gugger, Mariama Drame, Quentin Lhoest,
  and Alexander Rush. 2020.
\newblock \href {https://doi.org/10.18653/v1/2020.emnlp-demos.6} {Transformers:
  State-of-the-art natural language processing}.
\newblock In \emph{Proceedings of the 2020 Conference on Empirical Methods in
  Natural Language Processing: System Demonstrations}, pages 38--45, Online.
  Association for Computational Linguistics.

\bibitem[{Xie et~al.(2022)Xie, Shen, Li, Mao, and Han}]{xie2022eider}
Yiqing Xie, Jiaming Shen, Sha Li, Yuning Mao, and Jiawei Han. 2022.
\newblock \href {https://doi.org/10.18653/v1/2022.findings-acl.23} {Eider:
  Empowering document-level relation extraction with efficient evidence
  extraction and inference-stage fusion}.
\newblock In \emph{Findings of the Association for Computational Linguistics:
  ACL 2022}, pages 257--268, Dublin, Ireland. Association for Computational
  Linguistics.

\bibitem[{Xu et~al.(2021)Xu, Wang, Lyu, Zhu, and Mao}]{xu2021ssan}
Benfeng Xu, Quan Wang, Yajuan Lyu, Yong Zhu, and Zhendong Mao. 2021.
\newblock \href {https://arxiv.org/abs/2102.10249} {Entity structure within and
  throughout: Modeling mention dependencies for document-level relation
  extraction}.
\newblock In \emph{Proceedings of the AAAI conference on artificial
  intelligence}, volume~35, pages 14149--14157.

\bibitem[{Yang et~al.(2020)Yang, Zheng, Dai, He, Huang, and
  Chen}]{yang2020enhance}
Kaijia Yang, Nantao Zheng, Xinyu Dai, Liang He, Shujian Huang, and Jiajun Chen.
  2020.
\newblock \href {https://dl.acm.org/doi/abs/10.1145/3340531.3412153} {Enhance
  prototypical network with text descriptions for few-shot relation
  classification}.
\newblock In \emph{Proceedings of the 29th ACM International Conference on
  Information \& Knowledge Management}, pages 2273--2276.

\bibitem[{Yao et~al.(2019)Yao, Ye, Li, Han, Lin, Liu, Liu, Huang, Zhou, and
  Sun}]{yao2019docred}
Yuan Yao, Deming Ye, Peng Li, Xu~Han, Yankai Lin, Zhenghao Liu, Zhiyuan Liu,
  Lixin Huang, Jie Zhou, and Maosong Sun. 2019.
\newblock \href {https://doi.org/10.18653/v1/P19-1074} {{D}oc{RED}: A
  large-scale document-level relation extraction dataset}.
\newblock In \emph{Proceedings of the 57th Annual Meeting of the Association
  for Computational Linguistics}, pages 764--777, Florence, Italy. Association
  for Computational Linguistics.

\bibitem[{Ye et~al.(2020)Ye, Lin, Du, Liu, Li, Sun, and
  Liu}]{ye2020coreferential}
Deming Ye, Yankai Lin, Jiaju Du, Zhenghao Liu, Peng Li, Maosong Sun, and
  Zhiyuan Liu. 2020.
\newblock \href {https://doi.org/10.18653/v1/2020.emnlp-main.582}
  {{C}oreferential {R}easoning {L}earning for {L}anguage {R}epresentation}.
\newblock In \emph{Proceedings of the 2020 Conference on Empirical Methods in
  Natural Language Processing (EMNLP)}, pages 7170--7186, Online. Association
  for Computational Linguistics.

\bibitem[{Zaporojets et~al.(2021)Zaporojets, Deleu, Develder, and
  Demeester}]{zaporojets2021dwie}
Klim Zaporojets, Johannes Deleu, Chris Develder, and Thomas Demeester. 2021.
\newblock \href {https://arxiv.org/abs/2009.12626} {Dwie: An entity-centric
  dataset for multi-task document-level information extraction}.
\newblock \emph{Information Processing \& Management}, 58(4):102563.

\bibitem[{Zhang et~al.(2021)Zhang, Chen, Xie, Deng, Tan, Chen, Huang, Si, and
  Chen}]{zhang2021docunet}
Ningyu Zhang, Xiang Chen, Xin Xie, Shumin Deng, Chuanqi Tan, Mosha Chen, Fei
  Huang, Luo Si, and Huajun Chen. 2021.
\newblock \href {https://doi.org/10.24963/ijcai.2021/551} {Document-level
  relation extraction as semantic segmentation}.
\newblock In \emph{Proceedings of the Thirtieth International Joint Conference
  on Artificial Intelligence, {IJCAI-21}}, pages 3999--4006. International
  Joint Conferences on Artificial Intelligence Organization.
\newblock Main Track.

\bibitem[{Zhang and Lu(2022)}]{zhang2022better}
Peiyuan Zhang and Wei Lu. 2022.
\newblock \href {https://aclanthology.org/2022.emnlp-main.471} {Better few-shot
  relation extraction with label prompt dropout}.
\newblock In \emph{Proceedings of the 2022 Conference on Empirical Methods in
  Natural Language Processing}, pages 6996--7006, Abu Dhabi, United Arab
  Emirates. Association for Computational Linguistics.

\bibitem[{Zhang et~al.(2017)Zhang, Zhong, Chen, Angeli, and
  Manning}]{zhang2017position}
Yuhao Zhang, Victor Zhong, Danqi Chen, Gabor Angeli, and Christopher~D.
  Manning. 2017.
\newblock \href {https://doi.org/10.18653/v1/D17-1004} {Position-aware
  attention and supervised data improve slot filling}.
\newblock In \emph{Proceedings of the 2017 Conference on Empirical Methods in
  Natural Language Processing}, pages 35--45, Copenhagen, Denmark. Association
  for Computational Linguistics.

\bibitem[{Zhou et~al.(2021)Zhou, Huang, Ma, and Huang}]{zhou2021atlop}
Wenxuan Zhou, Kevin Huang, Tengyu Ma, and Jing Huang. 2021.
\newblock \href {https://arxiv.org/abs/2010.11304} {Document-level relation
  extraction with adaptive thresholding and localized context pooling}.
\newblock In \emph{Proceedings of the AAAI Conference on Artificial
  Intelligence}, volume~35, pages 14612--14620.

\bibitem[{Zhou et~al.(2023)Zhou, Zhang, Naumann, Chen, and
  Poon}]{zhou2022continual}
Wenxuan Zhou, Sheng Zhang, Tristan Naumann, Muhao Chen, and Hoifung Poon. 2023.
\newblock \href {https://doi.org/10.18653/v1/2023.acl-long.739} {Continual
  contrastive finetuning improves low-resource relation extraction}.
\newblock In \emph{Proceedings of the 61st Annual Meeting of the Association
  for Computational Linguistics (Volume 1: Long Papers)}, pages 13249--13263,
  Toronto, Canada. Association for Computational Linguistics.

\end{thebibliography}
\bibliographystyle{acl_natbib}


\appendix
\section*{Appendix}
\label{sec:appendix}

\section{Additional Dataset Details} \label{appendix:a}

\subsection{Data Statistics}

We report the statistics for the datasets in Table~\ref{tab:data_stats}. The test set of DocRED is not included in calculating \textbf{\% NA} due to its inaccessibility. 14 documents in DWIE are filtered out because of missing labels, and 1 document is removed because the annotated entities did not exist in the input document.

\begin{table}[h]
    \centering
    \begin{adjustbox}{width=\linewidth}
    {\small
    \begin{tabular}{l c c c}
    \noalign{\hrule height 1pt}
        Statistics & DocRED & Re-DocRED & DWIE \\
    \hline
        \# Train & 3,053 & 3,053 & 587 \\
        \# Dev & 1,000 & 500 & 100 \\
        \# Test & 1,000 & 500 & 100 \\
        \# Relation Types & 97 & 97 & 66 \\
        \% NA & 97.05\% & 94.02\% & 97.87\% \\
    \noalign{\hrule height 1pt}
    \end{tabular}
    }
    \end{adjustbox}
    \caption{Dataset statistics. \textbf{\# Relation Types} includes the \texttt{NA} label. \textbf{\% NA} indicates a ratio of entity pairs having the \texttt{NA} label over all entity pairs.}
    \label{tab:data_stats}
\end{table}

\subsection{Class Distribution}

\begin{figure}[h]
    \centering
    \includegraphics[width=\linewidth]{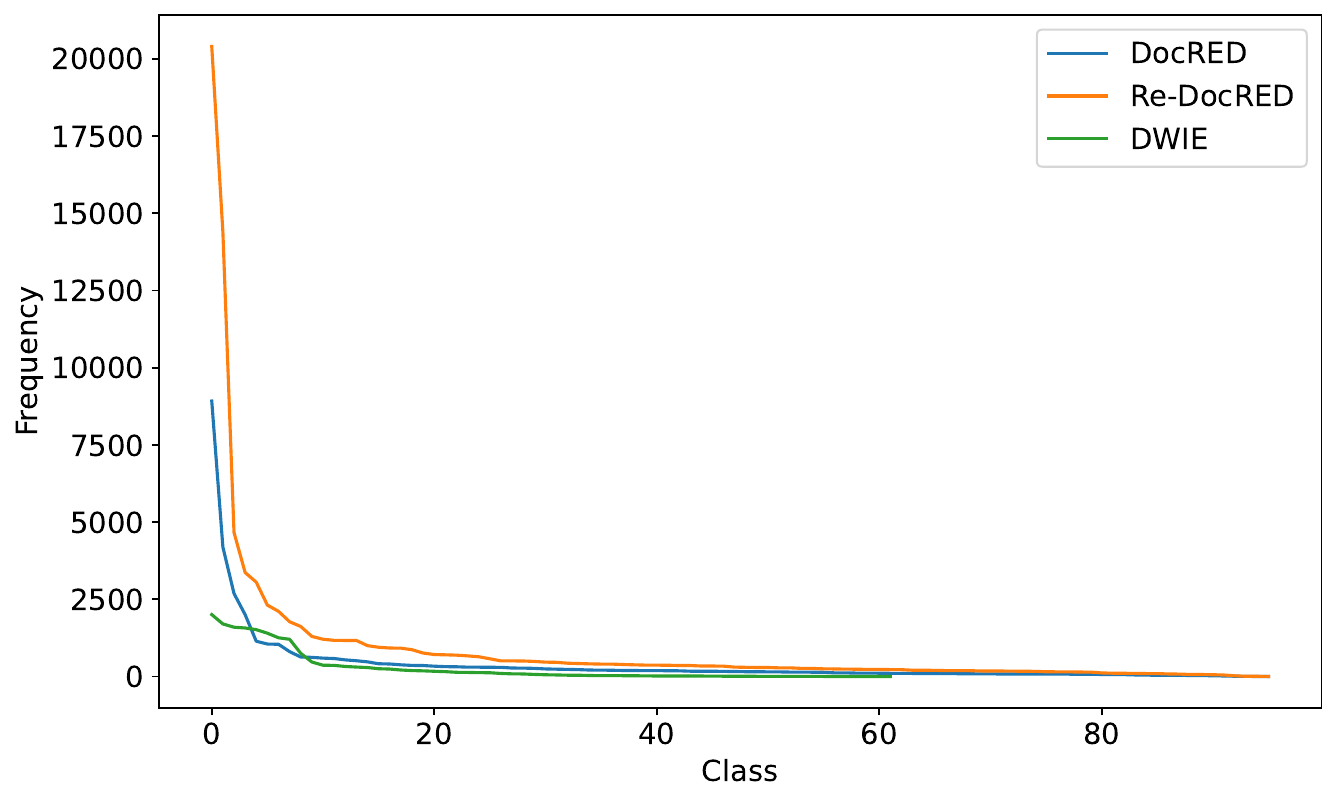}
    \caption{Dataset class distribution.}
    \label{fig:class_dist}
\end{figure}

We count the number of ground-truth relations in the train sets and visualize their class distribution in Figure~\ref{fig:class_dist}. We observe that the re-annotation of Re-DocRED further skewed the class distribution, and the DWIE dataset seems to demonstrate a relatively less imbalanced distribution. Nevertheless, a few classes still exhibit high frequency, which PRiSM can handle effectively.

\subsection{Processing Long Document} \label{appendix:a3}

\begin{figure}[h]
    \centering
    \includegraphics[width=\linewidth]{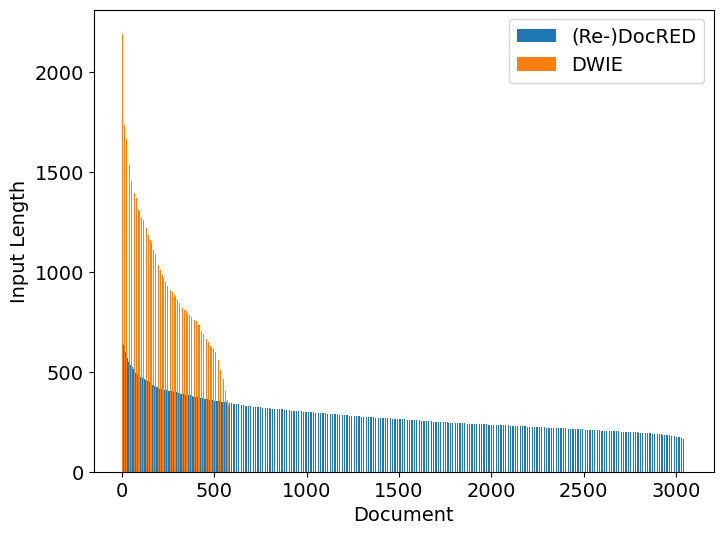}
    \caption{Lengths of input documents in training sets.}
    \label{fig:doc_length}
\end{figure}

For DocRED and Re-DocRED, most of the documents contain less than 512 tokens, and thus we follow the previous work and truncate all of the inputs to 512 tokens, which is the maximum sequence length of BERT. However, we notice that the DWIE dataset mostly contains documents much longer than 512 tokens (as shown in Figure~\ref{fig:doc_length}) in which the truncation hurts the performance significantly. Therefore, we choose the most naive way of splitting the input document into multiple chunks of length 512 and passing them through the encoder multiple times. The performance improvement over the truncation method is demonstrated in Table~\ref{tab:long_doc}.

\begin{table}[h]
    \centering
    \begin{adjustbox}{width=\linewidth}
    {\small
    \begin{tabular}{l c c}
        \noalign{\hrule height 1pt}
        \textbf{Method} & Dev $F_1$ & Test $F_1$ \\
        \hline
        \multicolumn{3}{l}{$N=100$} \\
        \hline
        Truncation & $36.53 \pm 2.09$ & $35.39 \pm 0.71$ \\
        Chunking (ours) & $\bm{45.20} \pm 1.60$ & $\bm{44.31} \pm 1.47$ \\
        \hline
        \multicolumn{3}{l}{$N=305$} \\
        \hline
        Truncation & $53.65 \pm 0.63$ & $51.30 \pm 0.62$ \\
        Chunking (ours) & $\bm{58.23} \pm 0.40$ & $\bm{57.05} \pm 0.23$ \\
        \hline
        \multicolumn{3}{l}{$N=587$} \\
        \hline
        Truncation & $61.34 \pm 0.52$ & $59.17 \pm 1.86$\\
        Chunking (ours) & $\bm{66.81} \pm 0.56$ & $\bm{66.53} \pm 0.52$ \\
    \noalign{\hrule height 1pt}
    \end{tabular}
    }
    \end{adjustbox}
    \caption{Performance comparison of long document processing methods. The model is fixed with BERT\textsubscript{BASE} + PRiSM evaluating on the DWIE dataset.}
    \label{tab:long_doc}
\end{table}

\subsection{Relation Descriptions} \label{appendix:a4}

We provide a small set of relations and their descriptions in DocRED and DWIE in Table~\ref{tab:reldesc_docred} and \ref{tab:reldesc_dwie}.
For DocRED, a full list can be found either in their paper~\cite{yao2019docred} or link\footnote{\url{https://www.wikidata.org/wiki/Wikidata:Main_Page}}.
For DWIE, a full list is not available publicly; however, we were able to obtain a draft of the annotation documentation from the author. Unannotated relation descriptions were crafted with the help of a large language model~\cite{openai2023chatgpt}.

\begin{table}[h]
    \centering
    \begin{adjustbox}{width=\linewidth}
    {\small
    \begin{tabular}{l p{.8\linewidth}}
        \noalign{\hrule height 1pt}
        \textbf{Relation Name} & \textbf{Description} \\
        \hline
        head of government & head of the executive power of this town, city, municipality, state, country, or other governmental body \\
        country & sovereign state of this item; don’t use on humans \\
        place of birth & most specific known (e.g. city instead of country, or hospital instead of city) birth location of a person, animal or fictional character \\
        country of citizenship & the object is a country that recognizes the subject as its citizen \\
        member of sports team & sports teams or clubs that the subject currently represents or formerly represented \\
        \noalign{\hrule height 1pt}
    \end{tabular}
    }
    \end{adjustbox}
    \caption{DocRED relation descriptions.}
    \label{tab:reldesc_docred}
\end{table}

\begin{table}[h]
    \centering
    \begin{adjustbox}{width=\linewidth}
    {\small
    \begin{tabular}{l p{.8\linewidth}}
        \noalign{\hrule height 1pt}
        \textbf{Relation Name} & \textbf{Description} \\
        \hline
        based\_in0 & Relations between organizations and the countries they are based in \\
        in0 & Relations between geographic locations and the countries they are located in \\
        citizen\_of & Relations between people and the country they are citizens of \\
        based\_in0-x & Relations between organizations and the nominal variations of the countries they are based in \\
        citizen\_of-x & Relations between people and the nominal variations of the countries they are citizens of \\
        \noalign{\hrule height 1pt}
    \end{tabular}
    }
    \end{adjustbox}
    \caption{DWIE relation descriptions.}
    \label{tab:reldesc_dwie}
\end{table}

\section{Additional Details for PRiSM} \label{appendix:b}

\subsection{Detailed Experimental Setup}

\paragraph{Device.}
For all our experiments, we trained the networks on a single NVIDIA TITAN RTX GPU with 24GB of memory.

\paragraph{Model Size.}
PRiSM shares parameters with the PLM used when learning relation representations. The only additional parameter weights come from a linear layer constructing pair representations and an extra embedding space initialized for relation tokens. The number of trainable parameters for each model is illustrated in Table~\ref{tab:model_size}.

\begin{table}[h]
    \centering
    \begin{adjustbox}{width=.9\linewidth}
    {\small
        \begin{tabular}{ l | c }
            \noalign{\hrule height 1pt}
            \textbf{Model} & \# Parameters \\
            \hline
            BERT\textsubscript{BASE} & 108,310,272 \\
            BERT\textsubscript{BASE}-DocRE & 114,259,297 \\
            BERT\textsubscript{BASE}-DocRE + \textbf{PRiSM} & 115,514,209 \\
            RoBERTa\textsubscript{BASE} & 124,645,632 \\
            RoBERTa\textsubscript{BASE}-DocRE & 130,594,657 \\
            RoBERTa\textsubscript{BASE}-DocRE + \textbf{PRiSM} & 131,849,569 \\
            \noalign{\hrule height 1pt}
        \end{tabular}
    }
    \end{adjustbox}
    \caption{Comparison of model parameters. \textbf{DocRE} includes a bilinear layer and two linear layers for constructing head and tail representations.}
    \label{tab:model_size}
\end{table}

\paragraph{GPU Hours.}

Adding PRiSM takes a slightly longer computation time than the existing DocRE models due to having to pass the PLM twice.
Note that PRiSM is built for a low-resource setting in which the computation time does not seem to differ as much. The comparison of GPU hours is reported in Table~\ref{tab:gpu_hours}.

\begin{table}[h]
    \centering
    \begin{adjustbox}{width=.7\linewidth}
    {\small
        \begin{tabular}{ l c }
            \noalign{\hrule height 1pt}
            \textbf{Model} & Training Hours \\
            \hline
            \multicolumn{2}{l}{\textit{3\% training examples}} \\
            \hline
            BERT\textsubscript{BASE} & 0.8 \\
            BERT\textsubscript{BASE} + \textbf{PRiSM} & 0.8 \\
            \hline
            \multicolumn{2}{l}{\textit{10\% training examples}} \\
            \hline
            BERT\textsubscript{BASE} & 0.8 \\
            BERT\textsubscript{BASE} + \textbf{PRiSM} & 1.0 \\
            \hline
            \multicolumn{2}{l}{\textit{100\% training examples}} \\
            \hline
            BERT\textsubscript{BASE} & 2.7 \\
            BERT\textsubscript{BASE} + \textbf{PRiSM} & 3.5 \\
            \noalign{\hrule height 1pt}
        \end{tabular}
    }
    \end{adjustbox}
    \caption{GPU hours for DocRED training. Time for evaluating and saving the model every epoch is included.}
    \label{tab:gpu_hours}
\end{table}

\paragraph{Hyperparameters.}

We perform a grid search on finding the best hyperparameter configuration and report the tuning range used for our experiments in Table~\ref{tab:hyperparams}.
The evaluation on the validation set is performed for every epoch and the tolerance increases by 1 when the validation $F_1$ is worse than the previous evaluation. The training stops early when the count reaches the max tolerance.

\begin{table}[h]
    \centering
    \begin{adjustbox}{width=\linewidth}
    {\small
        \begin{tabular}{p{.25\linewidth}|c|c|>{\centering\arraybackslash}p{.3\linewidth}|c}
            \noalign{\hrule height 1pt}
            \textbf{Model} & \textbf{Dataset} & \textbf{Hyperparameter} & \textbf{Range} & \textbf{Best} \\
            \hline
            \multirow{8}{.24\linewidth}{BERT\textsubscript{BASE}, RoBERTa\textsubscript{BASE}} & \multirow{8}{*}{(Re-)DocRED} & batch size & \{ 4, 8 \} & 4 \\
            & & learning rate & \{ 1e-5, 2e-5, 3e-5, 4e-5, 5e-5, 1e-4 \} & 3e-5 \\
            & & warmup ratio & \{ 0, 0.06, 0.1 \} & 0.06 \\
            & & $\lambda$ & \{ 1, 5, 10, 100 \} & 10 \\
            & & max grad norm & \{ 1.0 \} & 1.0 \\
            & & max tolerance & \{ 5 \} & 5 \\
            & & epoch & \{ 30 \} & 30 \\
            \hline
            \multirow{8}{.24\linewidth}{BERT\textsubscript{BASE}, RoBERTa\textsubscript{BASE}} & \multirow{8}{*}{DWIE} & batch size & \{ 4, 8 \} & 4 \\
            & & learning rate & \{ 1e-5, 2e-5, 3e-5, 4e-5, 5e-5, 1e-4 \} & 5e-5 \\
            & & warmup ratio & \{ 0, 0.06, 0.1 \} & 0.06 \\
            & & $\lambda$ & \{ 1, 5, 10, 100 \} & 10 \\
            & & max grad norm & \{ 1.0 \} & 1.0 \\
            & & max tolerance & \{ 5 \} & 5 \\
            & & epoch & \{ 30 \} & 30 \\
            \noalign{\hrule height 1pt}
        \end{tabular}
    }
    \end{adjustbox}
    \caption{Hyperparameter tuning range and best values used in the experiments.}
    \label{tab:hyperparams}
\end{table}

\begin{figure*}[hbt!]
    \centering
    \includegraphics[width=\textwidth]{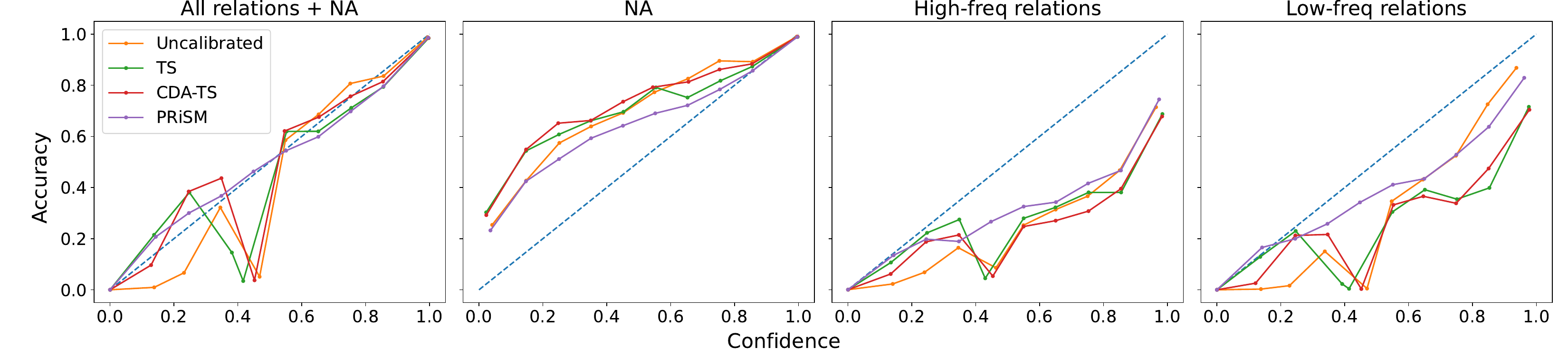}
    \caption{Reliability diagram for BERT\textsubscript{BASE} when trained with 10\% of DocRED data.}
    \label{fig:N305_calibration}
\end{figure*}

\begin{figure*}[hbt!]
    \centering
    \includegraphics[width=\textwidth]{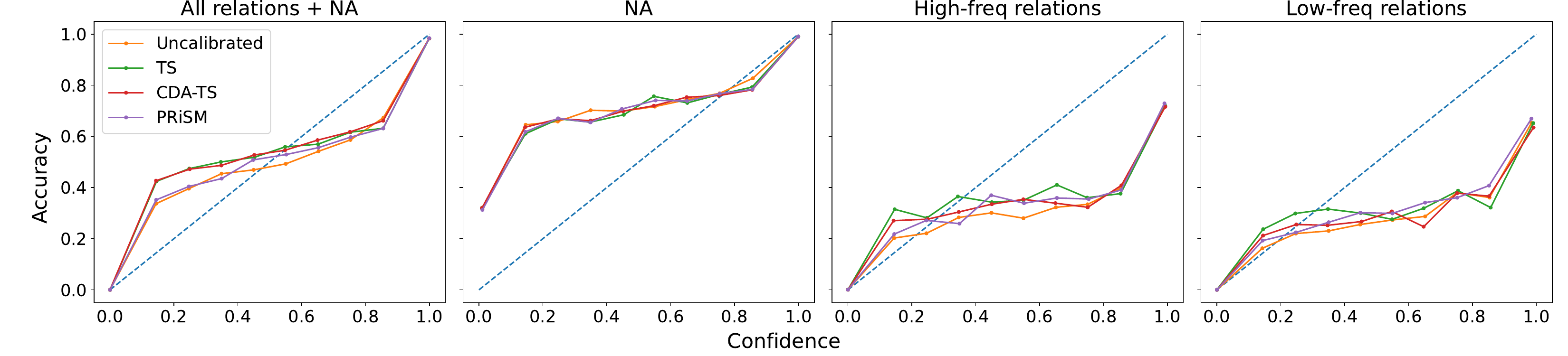}
    \caption{Reliability diagram for BERT\textsubscript{BASE} when trained with 100\% of DocRED data.}
    \label{fig:N3053_calibration}
\end{figure*}

\subsection{Evaluation Details}
We elaborate on the details of calculating calibration errors. We utilize two metrics in our paper. ECE~\cite{naeini2015bayesian} divides the probability interval into a fixed number of bins, calculates the difference between the accuracy of the predictions and the mean of the probabilities (confidence) in each bin, and computes a weighted average over the bins. Formally, the equation can be written as
\begin{equation}
    \text{ECE} = \sum_{b=1}^{B} \frac{n_b}{T}|\text{acc}(b)-\text{conf}(b)|,
\end{equation}
where $n_b$ is the number of predictions in bin $b$, $B$ is a hyperparameter for the total number of bins, and $T$ is the total number of samples.
On the other hand, ACE~\cite{nixon2019ace} divides up the probability interval by having the same number of predictions in each bin, thereby mitigating the issue of only calibrating the most confident samples.
The equation is written as
\begin{equation}
    \text{ACE} = \frac{1}{KR} \sum_{k=1}^{K} \sum_{r=1}^{R}|\text{acc}(r,k)-\text{conf}(r,k)|,
\end{equation}
where $\text{acc}(r,k)$ and $\text{conf}(r,k)$ are the accuracy and confidence of adaptive calibration range $r$ for class label $k$, respectively.

\subsection{Additional Calibration Results} \label{appendix:b3}

We visualize the calibration of the rest of the data setting (i.e., 10\% and 100\% training data) with reliability diagrams in Figure~\ref{fig:N305_calibration} and \ref{fig:N3053_calibration}. We notice that PRiSM is still effective with 10\% of training data, but with full data, the performance gain is minimal; that is, the line barely moves toward the perfect calibration line.

We understand that calibration results on models other than the BERT\textsubscript{BASE} may be important in demonstrating the effectiveness of PRiSM. As shown in Table~\ref{tab:calibration2}, we find that RoBERTa\textsubscript{BASE} and SSAN-BERT\textsubscript{BASE} follow the same trend as BERT\textsubscript{BASE}, showing the lowest calibration error when PRiSM is appended. We also observe a similar pattern with the Re-DocRED data. We do not report results for ATLOP because the calibration errors for ATLOP must be computed differently, as it does not use probabilities (confidence) for prediction.

\begin{table}[]
    \centering
    \begin{adjustbox}{width=\linewidth}
    {\small
        \begin{tabular}{ l | c c c | c c c }
            \noalign{\hrule height 1pt}
            & \multicolumn{3}{c|}{$N=100$} & \multicolumn{3}{c}{$N=305$} \\
            \hline
            \textbf{Method} & \bm{$F_1$}($\uparrow$) & \textbf{ECE}($\downarrow$) & \textbf{ACE}($\downarrow$) & \bm{$F_1$}($\uparrow$) & \textbf{ECE}($\downarrow$) & \textbf{ACE}($\downarrow$)\\
            \hline
            \multicolumn{7}{l}{\textit{DocRED results}} \\
            \hline
            BERT & 10.82 & 0.359\% & 0.379\% & 42.56 & 0.137\% & 0.164\% \\
            BERT + \textbf{PRiSM} & \textbf{37.84} & \textbf{0.010\%} & \textbf{0.020\%} & \textbf{48.10} & \textbf{0.023\%} & \textbf{0.020\%} \\
            RoBERTa & 22.55 & 0.671\% & 0.691\% & 45.83 & 0.237\% & 0.259\% \\
            RoBERTa + \textbf{PRiSM} & \textbf{35.10} & \textbf{0.015\%} & \textbf{0.025\%} & \textbf{48.70} & \textbf{0.022\%} & \textbf{0.020\%} \\
            SSAN-BERT & 11.93 & 0.368\% & 0.390\% & 42.82 & 0.128\% & 0.152\% \\
            SSAN-BERT + \textbf{PRiSM} & \textbf{36.96} & \textbf{0.019\%} & \textbf{0.019\%} & \textbf{48.28} & \textbf{0.023\%} & \textbf{0.023\%} \\
            \hline
            \multicolumn{7}{l}{\textit{Re-DocRED results}} \\
            \hline
            BERT & 32.75 & 0.367\% & 0.407\% & 54.44 & 0.185\% & 0.190\% \\
            BERT + \textbf{PRiSM} & \textbf{49.90} & \textbf{0.038\%} & \textbf{0.048\%} & \textbf{60.17} & \textbf{0.056\%} & \textbf{0.036\%} \\
            RoBERTa & 42.40 & 0.722\% & 0.718\% & 57.61 & 0.191\% & 0.208\% \\
            RoBERTa + \textbf{PRiSM} & \textbf{50.34} & \textbf{0.055\%} & \textbf{0.051\%} & \textbf{61.55} & \textbf{0.047\%} & \textbf{0.033\%} \\
            SSAN-BERT & 30.28 & 0.362\% & 0.407\% & 55.60 & 0.125\% & 0.148\% \\
            SSAN-BERT + \textbf{PRiSM} & \textbf{49.04} & \textbf{0.050\%} & \textbf{0.053\%} & \textbf{60.57} & \textbf{0.051\%} & \textbf{0.036\%} \\
            \noalign{\hrule height 1pt}
        \end{tabular}
    }
    \end{adjustbox}
    \caption{Comparison of calibration errors (with 10 bins) of different models under a low-resource setting of DocRED and Re-DocRED.}
    \label{tab:calibration2}
\end{table}

\end{document}